\pdfoutput=1

\documentclass[11pt]{article}

\usepackage[preprint]{acl}

\usepackage{times}
\usepackage{latexsym}

\usepackage[T1]{fontenc}

\usepackage[utf8]{inputenc}

\usepackage{microtype}

\usepackage{inconsolata}

\usepackage{graphicx}

\usepackage{amsmath} 
\usepackage{multirow}
\usepackage{multicol}
\usepackage{booktabs}  % Add this in your preamble
\usepackage{tabularx}
\usepackage{geometry}
\usepackage{enumitem}
\usepackage{subcaption}
\usepackage{amssymb}
\usepackage{makecell}
\usepackage[most]{tcolorbox}

\title{Benchmarking and Mitigating MCQA Selection Bias of Large Vision-Language Models}

\author{Md. Atabuzzaman \qquad Ali Asgarov\thanks{The proposed bias mitigation method was fully created and implemented by this author.}
\qquad Chris Thomas \\
Department of Computer Science\\
  Virginia Tech \\
  \texttt{\{atabuzzaman, aliasgarov, christhomas\}@vt.edu}\\}
\begin{document}
\maketitle
\begin{abstract}

Large Vision-Language Models (LVLMs) have achieved strong performance on vision-language tasks, particularly Visual Question Answering (VQA). While prior work has explored unimodal biases in VQA, the problem of selection bias in Multiple-Choice Question Answering (MCQA), where models may favor specific option tokens (e.g., "A") or positions, remains underexplored. In this paper, we investigate both the presence and nature of selection bias in LVLMs through fine-grained MCQA benchmarks spanning easy, medium, and hard difficulty levels, defined by the semantic similarity of the options. We further propose an inference-time logit-level debiasing method that estimates an ensemble bias vector from general and contextual prompts and applies confidence-adaptive corrections to the model’s output. Our method mitigates bias without retraining and is compatible with frozen LVLMs. Extensive experiments across several state-of-the-art models reveal consistent selection biases that intensify with task difficulty, and show that our mitigation approach significantly reduces bias while improving accuracy in challenging settings. This work offers new insights into the limitations of LVLMs in MCQA and presents a practical approach to improve their robustness in fine-grained visual reasoning. Datasets and code are available at: \url{https://github.com/Atabuzzaman/Selection-Bias-of-LVLMs}

\end{abstract}

\begin{figure}[!t]
    \centering
    
    \begin{subfigure}[t]{\linewidth}
        \centering
        \includegraphics[width=\linewidth]{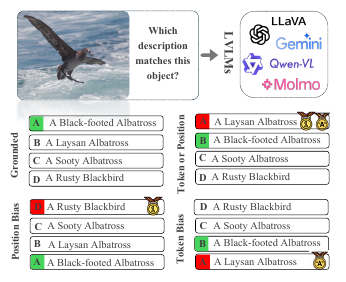}
        \vspace{-0.75cm}
        \caption{Illustration of selection bias (positional and token identity) in LVLM predictions (\textcolor{red}{red}) for visual multiple-choice question answering. Answer preferences change with option order and token labels. Correct answers are in \textcolor{green}{green}.}
        \vspace{0.2cm}
    \end{subfigure}
    
    \begin{subfigure}[t]{\linewidth}
        \centering
        \includegraphics[width=\linewidth]{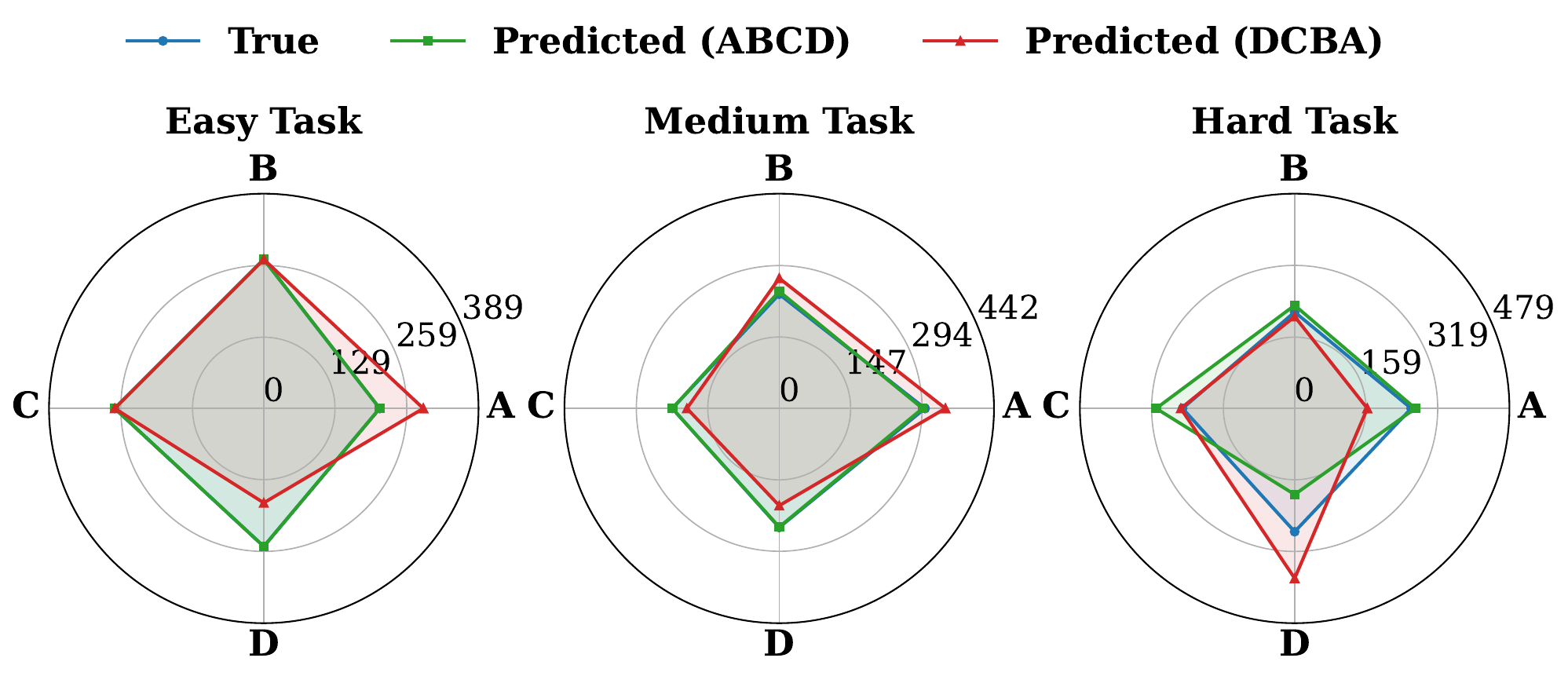}
        \vspace{-0.6cm}
        \caption{As task difficulty increases, Qwen2.5-VL-3B-Instruct exhibits stronger selection biases favoring option ID "A" in easy tasks (\textbf{token bias}) and first-position options "A" and "D" in "ABCD" and "DCBA" orderings, respectively, in hard tasks (\textbf{positional bias}).}
        \label{fig:qwen_3b}
    \end{subfigure}
    \vspace{-0.3cm}
    \caption{(a) Top: Visual illustration of selection bias in LVLMs. (b) Bottom: Amplification of token and positional biases in the Qwen2.5-VL-3B-Instruct model across increasing task difficulty.}
    \label{fig:teaser_combined}
    \vspace{-0.7cm}
\end{figure}

\begin{figure*}[t]
    \centering
    \includegraphics[width=\linewidth]{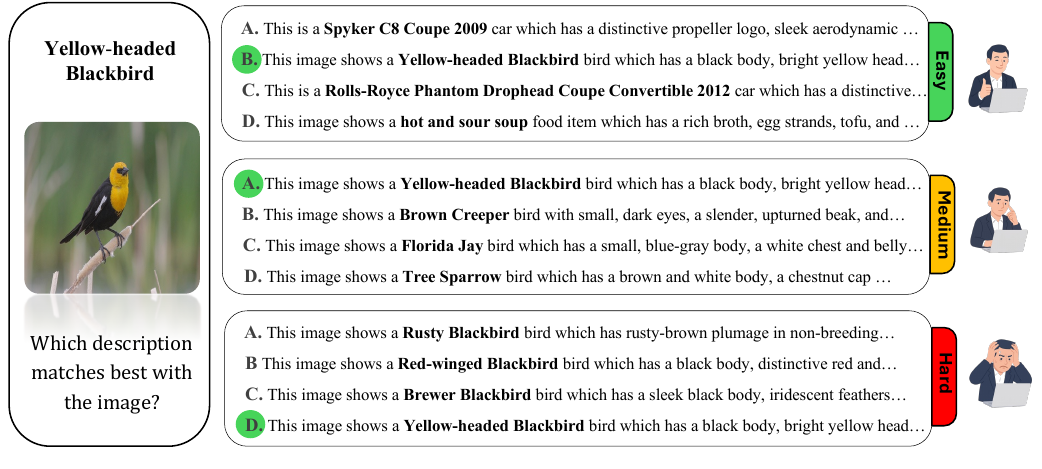}
    \vspace{-0.9cm}
    \caption{Examples from our fine-grained visual multiple-choice question answering benchmark for the "Yellow-headed Blackbird" class across three difficulty levels: easy, medium, and hard. Each example presents a multiple-choice question requiring the model to match an image with the most appropriate textual description. The easy task includes distractors (incorrect options) from different domains (e.g., vehicles, food), making the correct choice easily distinguishable. The medium task increases difficulty by using distractors from the same domain (i.e., birds) with less similar visual characteristics. The hard task presents the most visually similar bird species (e.g., blackbirds with subtle distinctions), demanding fine-grained reasoning. This structured difficulty progression enables systematic evaluation of LVLMs' reasoning capabilities and their susceptibility to selection biases, especially when class names are explicitly included or excluded in the options.}
    \vspace{-0.5cm}
  \label{fig:dataset_ex}
\end{figure*}

%%%%%%%%%%%%%%%%%%%%%%%%%%%%%%%%%%%%%%%%%%%%%%%%%%%%%%%%%%

\section{Introduction}

Large Vision-Language Models (LVLMs) \citep{llava2, llava, blip2, instructblip, molmo, Qwen2.5-VL, internvl2_5, gpt4o, januspro} have achieved impressive performance across a wide range of multimodal tasks, including visual question answering (VQA), image captioning, and visual reasoning. These models exhibit strong zero-shot generalization, attributed to pretraining on large-scale vision-language corpora and subsequent instruction tuning. However, despite their overall success, recent studies have revealed that LVLMs, like their text-only counterparts, are prone to various forms of bias that compromise the fairness, interpretability, and robustness of their outputs~\cite{adila2024discovering}.

One such underexplored phenomenon is selection bias in MCQA. Unlike open-ended VQA formats, MCQA requires models to select one option among predefined choices (e.g., A/B/C/D), introducing the possibility of preference for certain option positions or tokens. Similar forms of bias, such as position bias and token prior bias, have been documented in large language models (LLMs) \cite{pezeshkpour2024large, zheng2023judging}, but their manifestation in LVLMs remains relatively unexamined. Our preliminary findings suggest that LVLMs exhibit consistent preferences for specific options (e.g., choosing "A" or "D" disproportionately), especially in scenarios where answer candidates are semantically or visually similar (Figure \ref{fig:teaser_combined}). This can result in unstable or unreliable predictions that are influenced more by formatting than content \cite{adila2024discovering, zong2024fool}.

In this paper, we investigate the presence and nature of MCQA selection bias in LVLMs. We identify multiple sources of bias, including option position bias and token-level prior heuristics that models rely on instead of grounded visual reasoning. To study these phenomena in depth, we introduce new benchmark datasets designed specifically to evaluate LVLMs' MCQA selection behavior. Our dataset is constructed from fine-grained visual classification tasks and includes three levels of difficulty (Easy, Medium, and Hard) based on the semantic similarity between the correct option and distractor (or incorrect) options (Figure \ref{fig:dataset_ex}). We also incorporate variations with and without class names in the options to test the LVLM’s reliance on surface-level cues and prior domain knowledge.

To mitigate the selection bias, we propose an inference-time logit correction mechanism that adjusts the model’s output distribution over MCQ options based on an empirically estimated bias vector. Our method does not require retraining and is fully compatible with frozen pretrained LVLMs. It constructs an ensemble bias vector from both general prompts (capturing structural biases) and contextual prompts (reflecting task-specific tendencies), and adaptively corrects the model’s logits at inference based on prediction confidence. This approach counteracts option-token and positional biases while preserving the model’s ability to reason over visual and semantic content.

Through extensive experiments on several state-of-the-art (SOTA) LVLMs, we show that our fine-grained MCQA benchmarks reveal consistent and intensifying selection biases, particularly when visual evidence is inconclusive and options are fine-grained. We further demonstrate that our logit-based debiasing method improves model accuracy in challenging settings, enhances answer consistency under option reordering, and reduces reliance on spurious token and position priors.

Our main contributions are as follows:
\begin{itemize}[noitemsep, topsep=0pt]
    \item We propose benchmark datasets to study the selection bias of LVLMs across three difficulty level tasks: easy, medium, and hard.
    \item  Our datasets feature options with and without class names, revealing the nature of selection bias at each difficulty level and allowing investigation of LVLM behavior with and without prior domain knowledge.
    \item We propose an inference-time logit debiasing method that mitigates selection bias by correcting biased option-token distributions using an ensemble bias vector and confidence-adaptive scaling.
\end{itemize}

%%%%%%%%%%%%%%%%%%%%%%%%%%%%%%%%%%%%%%%%%%%%%%%%%%%%%%%%%

\section{Related Work} 
Several studies have explored the selection bias of LLMs~\citep{robinsonleveraging, zheng2023judging, zheng2024large, pezeshkpour2024large, xue2024strengthened, shi2024judging, balepur2025these, wang2025llms}. \citet{robinsonleveraging} showed that LLMs behave differently when prompted with option IDs compared to cloze-style prompts without IDs. They also evaluated the effect of varying the position of option IDs on model performance. \citet{zheng2023judging} \& \citet{wang2024large} found that GPT-4 tends to favor the first-presented answers, potentially leading to unfair evaluation outcomes. \citet{xue2024strengthened} argued that selection bias arises from the model's inability to effectively associate option IDs with the corresponding option text. \citet{pezeshkpour2024large} observed that LLMs are sensitive to changes in option order in MCQs and attributed this to positional bias and uncertainty. \citet{li2024can} highlighted selection bias in knowledge-intensive scenarios where long-form generation (LFG) is required. \citet{zheng2024large} introduced PriDe and demonstrated that removing option IDs shifts the main source of bias to the model's prior token bias, which can be mitigated through targeted debiasing. \citet{yang2024mitigating} addressed bias by removing neurons responsible for biased behavior. \citet{zhou2024unibias} proposed UniBias, an inference-only approach that identifies and eliminates biased feed-forward network (FFN) vectors and attention heads. \citet{wei2024unveiling} quantified the effects of token and option order on selection bias and mitigated them through weight and probability calibration. \citet{gudatiny} introduced a majority-voting method that reduces computational overhead while maintaining effective bias mitigation. Finally, \citet{yang2025option} proposed a causal debiasing technique that steers key component activations toward unbiased directions, applying stronger interventions to components with higher causal influence.

Recently, researchers have addressed various types of biases in Large Vision-Language Models (LVLMs) using techniques such as data augmentation~\citep{gokhale2020mutant}, model editing~\citep{cheng2023can, wang2024can}, and post-processing of outputs~\citep{wang2021gender, zhang2024debiasing}. \citet{zhang2024debiasing} proposed two strategies to mitigate language prior bias in classification tasks through output probability calibration. \citet{chen2024quantifying} assessed and mitigated LVLM bias in the VQA task by intervening in both questions and images using causal graphs. \citet{chen2024mitigating} proposed DCVC, a trainable output calibration network with virtual counterfactual augmentation, to reduce language bias in social intelligence QA. \citet{tan2024order} found that multimodal LLMs favor content at the beginning and end of contexts, and improved inference by strategically placing key elements. \citet{zong2024fool} identified permutation vulnerabilities in LVLM-based MCQs such as position bias and weak option-content links and introduced mitigation techniques like majority voting, confidence voting, and context calibration. \citet{adila2024discovering} proposed inference-type activation steering to reduce selection bias in LVLMs. However, effectively mitigating MCQA selection bias using pretrained LVLMs while preserving model capabilities and performance remains largely unexplored. In particular, how such biases evolve across tasks of increasing difficulty, from easy to medium to hard, has not been systematically studied or addressed.

\begin{figure}[!ht]
    \centering
    \begin{subfigure}[b]{\linewidth}
        \centering
        \includegraphics[width=\textwidth]{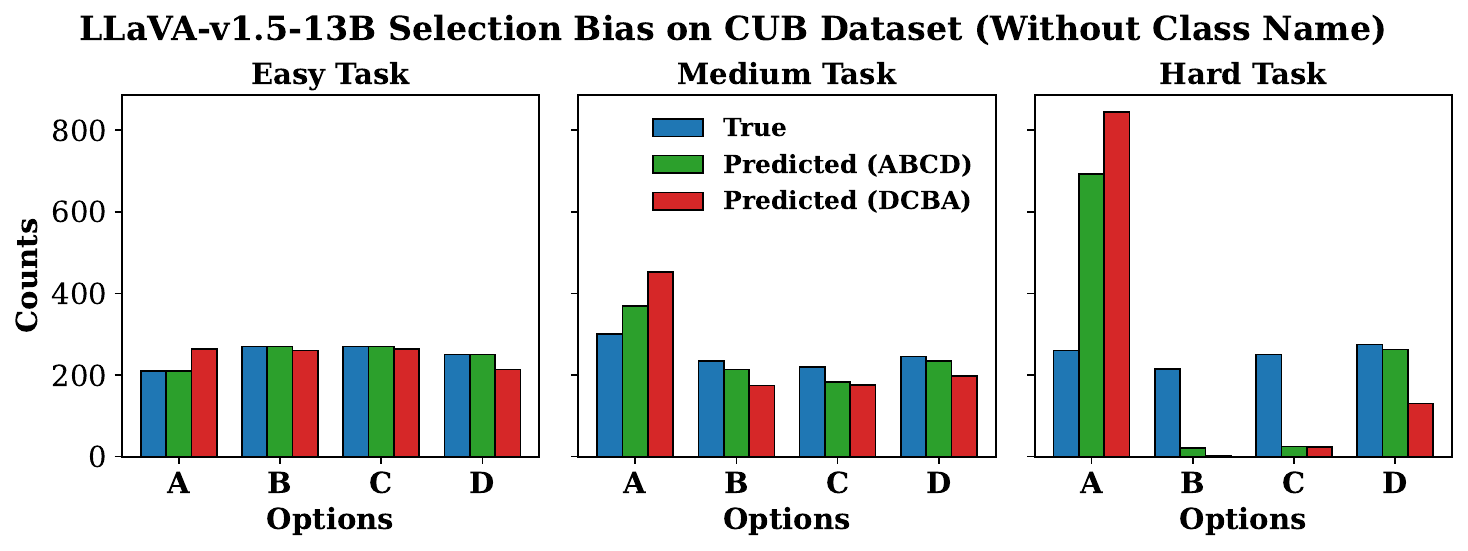}
        \vspace{-0.6cm}
        \caption{The LLaVA-v1.5-13B model shows balanced selection in easy tasks but develops strong token bias in hard tasks, with dramatic preference for option ID "A" (reaching nearly 3x the true frequency) when difficulty increases.}
        \label{fig:llava_13b_bias} 
        \vspace{-0.5cm}
    \end{subfigure}
    \vskip\baselineskip
    \begin{subfigure}[b]{\linewidth}
        \centering
        \includegraphics[width=\textwidth]{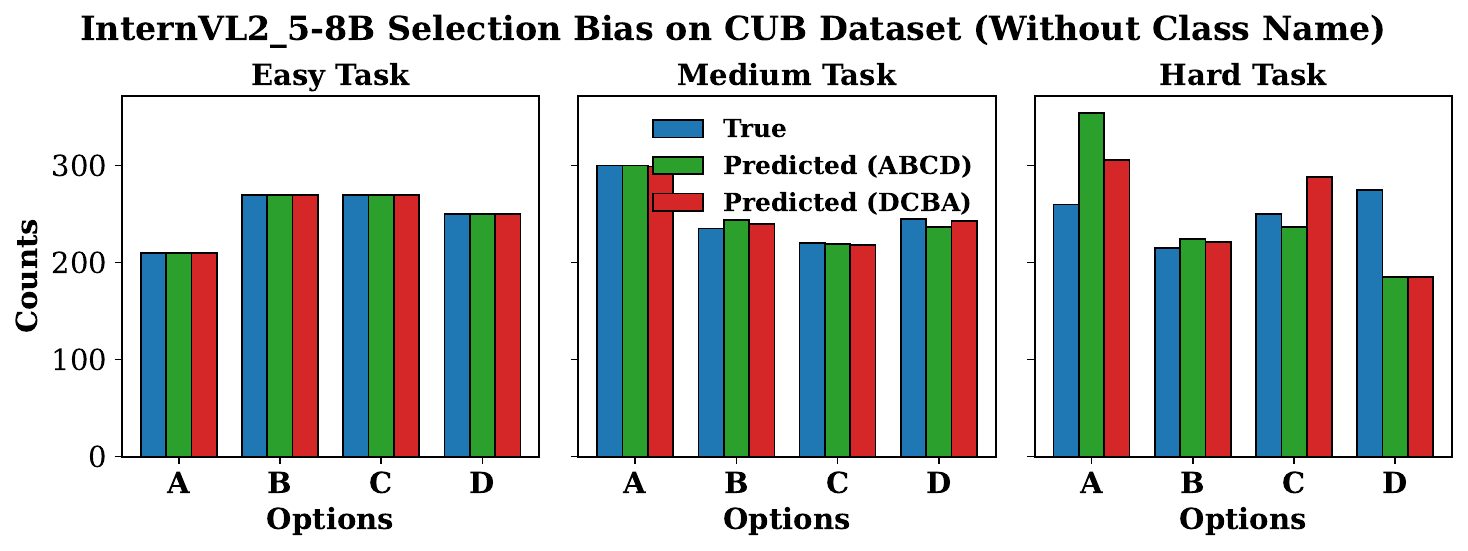}
        \vspace{-0.6cm}
        \caption{The InternVL2\_5-8B model demonstrates balanced behavior across difficulty levels, with predictions closely matching the true distribution in easy and medium tasks. In hard tasks, it shows moderate token bias toward "A" but maintains better distribution consistency between ABCD and DCBA orderings.}
        \label{fig:internvl_8b_bias}
    \end{subfigure}
    \vspace{-0.6cm}
    \caption{Selection bias comparison across two LVLMs on the CUB dataset under the "without class name" setting, organized by increasing task difficulty (easy, medium, hard). Each plot shows distributions for ground truth (True) and model predictions under two option orderings: standard (ABCD) and reversed (DCBA). The comparison reveals how position and token biases emerge and intensify with task difficulty, with varying patterns across architectures.}
    \label{fig:all_models_bias}
    \vspace{-0.5cm}
\end{figure}

%%%%%%%%%%%%%%%%%%%%%%%%%%%%%%%%%%%%%%%%%%%%%%%%%%%%%%%%
\section{Dataset Curation}
To evaluate the selection bias of LVLMs in the MCQA setting, we curate benchmark datasets that systematically test models under varying degrees of semantic similarity and domain familiarity. Our goal is to assess how effectively these models can select correct descriptive class text when presented with highly similar distractors (incorrect options), and to what extent their predictions are influenced by spurious correlations or prior knowledge.

Each MCQ in our dataset includes one correct class description and three incorrect class descriptions (distractors) as options, selected based on their cosine similarity to the correct description using CLIP's text encoder~\cite{clip}. By controlling this similarity, we categorize each question into one of three difficulty levels:

\textit{\textbf{Easy:}} Distractors are drawn from unrelated or semantically dissimilar classes. The differences between the correct option and the distractors are clear, even without domain-specific knowledge.

\textit{\textbf{Medium:}} Distractors are from the same domain but moderately different from the target class. These options require more reasoning to eliminate.

\textit{\textbf{Hard:}} Distractors are highly similar to the target class in terms of textual and semantic content, making them challenging to differentiate. These examples often require fine-grained understanding and detailed visual-textual alignment.

To further analyze how LVLM bias behavior changes when models can or cannot rely on domain knowledge, we create two versions of each MCQ:

\textit{\textbf{With Class Name:}} Class names are explicitly mentioned in the options. This version tests whether the model can leverage prior domain knowledge to reduce selection bias and improve prediction accuracy.

\textit{\textbf{Without Class Name:}} Class names are removed from the options, forcing the model to rely on fine-grained visual grounding and descriptive reasoning rather than recalling known labels, potentially increasing susceptibility to selection bias.

To ensure diverse coverage across semantic categories, we construct our benchmark using six fine-grained classification datasets: CUB-200-2011 \cite{cub} (200 bird species), Stanford Dogs \cite{dogs} (120 dog breeds), FGVC Aircraft \cite{aircrafts} (70/102 aircraft variants), Stanford Cars \cite{cars} (196 car models), Food-101 \cite{food} (101 food categories; referred to as Food-101 or Food throughout this paper), and iNaturalist-2021 \cite{di2021observing} (9962/10,000 species-level classes). Class descriptions are collected from \citet{atabuzzaman2025, finer}.

Using these class descriptions, for each image-class pair, we construct three difficulty-specific MCQs (easy, medium, hard) with two variants each, one with the class name included and one without. This results in six unique MCQs per image. For example, a dataset with 200 (e.g., CUB) classes would yield $200 \times 3 \times 2 = 1{,}200$ MCQs. \textbf{In total, our dataset contains 63,894 MCQs across 10,649 diverse classes.}

To validate our difficulty categorization, we compute the average and standard deviation of cosine similarity scores between correct and incorrect options across all difficulty levels. As expected, easy tasks show low similarity, medium tasks show moderate similarity, and hard tasks contain high-similarity incorrect options. Table~\ref{tab:similarity_stats} (Appendix~\ref{dataset_sta}) reports these statistics, confirming that our dataset reliably separates difficulty levels based on semantic similarity.

To enable effective detection of positional bias, we balance the position of the correct answer (A–D) across MCQs. Table~\ref{tab:answer_distribution} (Appendix~\ref{dataset_sta}) shows that all datasets maintain balanced correct answer position distributions across both settings (with and without class names), ensuring that any observed positional preferences reflect model bias rather than dataset imbalance. 

%%%%%%%%%%%%%%%%%%%%%%%%%%%%%%%%%%%%%%%%%%%%%%%%%%%%%%%%%

\section{Selection Bias of LVLMs}

In this section, we investigate the presence and nature of selection bias in LVLMs when applied to MCQA tasks. Our analysis focuses on three leading models: LLaVA-v1.5-13B~\cite{llava}, InternVL2\_5-8B~\cite{internvl2_5}, and Qwen2.5-VL-3B-Instruct~\cite{Qwen2.5-VL}. We identify consistent and model-specific selection biases that emerge during visual reasoning, particularly as task difficulty increases.

To disentangle token identity bias from positional bias, we design a comparative evaluation using both standard ("ABCD") and reversed ("DCBA") option orderings. As shown in Figures~\ref{fig:teaser_combined} and \ref{fig:all_models_bias}, all models exhibit relatively balanced option selection in easy tasks, with predicted distributions closely matching the ground-truth answer distribution. However, as difficulty increases to medium and hard levels, distinct and intensified bias patterns emerge. These patterns suggest that models begin to rely more heavily on heuristic behaviors, such as defaulting to specific token IDs—under semantic ambiguity, where fine-grained descriptions make it harder to distinguish between closely related classes.

\textbf{Token Bias Intensifies with Task Difficulty in LLaVA-v1.5-13B.} Figure~\ref{fig:llava_13b_bias} shows that LLaVA-v1.5-13B exhibits strong token bias in hard tasks, with the "A" option being selected nearly three times more often than expected. Crucially, this preference persists regardless of the option's actual position, indicating that the model defaults to the "A" token ID when uncertain. This behavior suggests the presence of a learned token-based prior that increasingly influences predictions as the model struggles with fine-grained visual-textual alignment. While the bias is negligible in easy tasks, it becomes dominant in hard tasks.

\textbf{InternVL2\_5-8B Exhibits More Balanced Token Selection Under Difficulty.} Figure~\ref{fig:internvl_8b_bias} demonstrates that InternVL2\_5-8B maintains more balanced token selection behavior across all levels of difficulty. Even in hard tasks where other models exhibit strong biases, InternVL2\_5-8B's predictions remain relatively aligned with the true answer distribution. Moreover, the model shows consistent behavior across both standard and reversed option formats, suggesting more robust reasoning and less susceptibility to token or positional artifacts. Nevertheless, InternVL still exhibits slight token ("A") bias amplification under difficulty, indicating that even stronger models fall back on heuristics when tasks become challenging.

\textbf{Qwen2.5-VL-3B-Instruct Reveals Complex Interaction Between Token and Positional Bias.} Figure~\ref{fig:qwen_3b} reveals that Qwen2.5-VL-3B-Instruct presents the most intricate bias behavior. The model exhibits a strong interaction between token identity and option position, particularly in hard tasks. In easy and medium tasks, Qwen2.5-VL-3B-Instruct shows mild bias toward "A" even when it appears last. However, in hard tasks, the bias pattern changes significantly: option "C" in the ABCD format becomes dominant, while option "D" when placed first in the DCBA format receives excessive selection. This suggests a conflation of positional and token-specific preferences. The interaction indicates the presence of multiple, competing biases within the model's decision-making process. The severe collapse to specific options under fine-grained semantic ambiguity implies heightened sensitivity to both token identity and position when semantic distinctions between options are subtle.

Our findings highlight important limitations in current LVLMs, particularly in fine-grained reasoning under semantic ambiguity. The variation in bias patterns across architectures suggests that each model encodes different shortcuts or priors. Furthermore, the consistent amplification of bias with increasing task difficulty aligns with observations from the LLM literature~\cite{pezeshkpour2024large}, where uncertainty leads models to default to preferred positions. \textbf{In our case, LVLMs exhibit both token and positional selection biases, revealing critical challenges in their reasoning reliability when fine-grained visual reasoning is required and MCQ options are semantically similar and fine-grained.}

%%%%%%%%%%%%%%%%%%%%%%%%%%%%%%%%%%%%%%%%%%%%%%%%%%%%
\begin{figure*}[htbp]
    \centering
    \includegraphics[width=\linewidth]{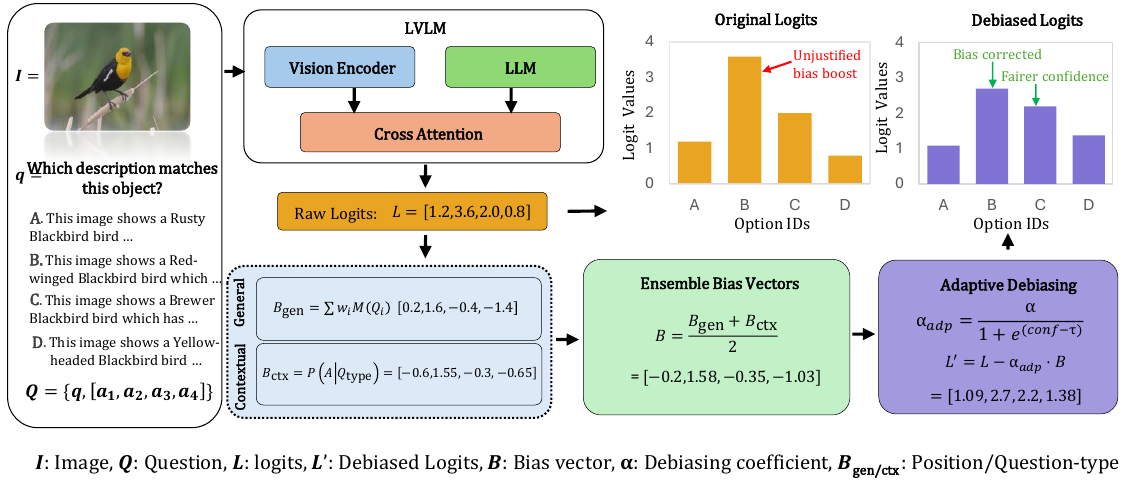}
    \vspace{-0.7cm}
    \caption{Illustration of our training-free ensemble debiasing framework for LVLM-based MCQA. We estimate general and contextual bias vectors, average them, and apply adaptive logit correction based on model confidence to reduce selection bias and improve prediction accuracy.}
    \label{fig:enter-label}
    \vspace{-0.5cm}
\end{figure*}

%%%%%%%%%%%%%%%%%%%%%%%%%%%%%%%%%%%%%%%%%%%%%%%%%%%
\begin{table*}[ht]
\centering
\small
\begin{tabular}{llcccc}
\toprule
\textbf{Class Name} & \textbf{Dataset (Difficulty)} & \textbf{Qwen2.5-VL-7B} & \textbf{Qwen2.5-VL-3B} & \textbf{InternVL2\_5-8B} & \textbf{LLaVA-v1.5-13B} \\
\midrule

\multirow{20}{*}{\textbf{With}} 
& Aircraft (Easy)   & 100.0 (100.0) & 100.0 (83.71) & 100.0 (100.0) & 100.0 (91.43) \\
& Aircraft (Medium) & 99.71 (100.0) & 98.00 (43.14) & 95.71 (95.43) & 48.86 (42.29) \\
& Aircraft (Hard)   & 71.14 (71.43) & 65.71 (37.71) & 51.14 (50.86) & 21.71 (23.71) \\ \cmidrule(lr){2-6}
& Cars (Easy)        & 100.0 (100.0) & 100.0 (81.43) & 100.0 (100.0) & 100.0 (94.08) \\
& Cars (Medium)      & 99.90 (99.59) & 99.69 (85.41) & 99.59 (99.49) & 84.69 (68.98) \\
& Cars (Hard)        & 82.14 (80.51) & 76.84 (55.71) & 57.04 (58.57) & 29.90 (35.51) \\ \cmidrule(lr){2-6}
& CUB (Easy)        & 100.0 (99.90) & 100.0 (93.40) & 100.0 (100.0) & 99.70 (86.40) \\
& CUB (Medium)      & 99.90 (99.70) & 99.70 (84.90) & 100.0 (99.50) & 62.70 (51.50) \\
& CUB (Hard)        & 76.70 (72.10) & 69.60 (48.00) & 60.50 (66.00) & 28.30 (33.70) \\ \cmidrule(lr){2-6}
& Dogs (Easy)       & 100.0 (100.0) & 100.0 (73.17) & 100.0 (100.0) & 98.50 (68.33) \\
& Dogs (Medium)     & 98.50 (98.67) & 96.83 (67.33) & 96.67 (99.17) & 45.00 (44.17) \\
& Dogs (Hard)       & 76.83 (77.50) & 71.00 (49.83) & 62.50 (55.83) & 29.17 (31.67) \\ \cmidrule(lr){2-6}
& Food (Easy)       & 100.0 (99.01) & 100.0 (79.41) & 100.0 (100.0) & 99.80 (91.29) \\
& Food (Medium)     & 98.22 (97.62) & 98.02 (72.87) & 98.22 (98.61) & 89.90 (77.62) \\
& Food (Hard)       & 89.11 (85.94) & 83.17 (58.42) & 84.55 (81.39) & 54.46 (58.42) \\ \cmidrule(lr){2-6}
& iNaturalist (Easy)       & 98.88 (95.18) & 97.99 (76.56) & 98.83 (98.59) & 68.61 (58.36) \\
& iNaturalist (Medium)     & 87.95 (85.87) & 84.11 (60.45) & 84.38 (84.56) & 45.44 (40.40) \\
& iNaturalist (Hard)       & 47.29 (44.87) & 41.91 (31.88) & 25.06 (40.87) & 25.46 (27.04) \\
\midrule

\multirow{20}{*}{\textbf{Without}} 
& Aircraft (Easy)   & 100.0 (100.0) & 100.0 (86.29) & 100.0 (100.0) & 100.0 (99.43) \\
& Aircraft (Medium) & 84.57 (79.43) & 80.57 (42.57) & 69.43 (75.71) & 38.57 (33.43) \\
& Aircraft (Hard)   & 53.71 (52.86) & 45.14 (33.71) & 40.29 (45.14) & 38.86 (36.29) \\ \cmidrule(lr){2-6}
& Cars (Easy)        & 100.0 (100.0) & 100.0 (79.90) & 100.0 (100.0) & 99.90 (98.98) \\
& Cars (Medium)      & 99.49 (98.47) & 99.49 (74.80) & 95.41 (96.84) & 84.49 (66.73) \\
& Cars (Hard)        & 70.00 (67.86) & 63.37 (44.49) & 54.69 (56.43) & 40.41 (35.61) \\ \cmidrule(lr){2-6}
& CUB (Easy)        & 100.0 (100.0) & 100.0 (92.60) & 100.0 (100.0) & 100.0 (94.70) \\
& CUB (Medium)      & 99.40 (99.00) & 99.10 (88.60) & 98.70 (99.10) & 91.80 (82.80) \\
& CUB (Hard)        & 64.90 (63.30) & 61.80 (46.40) & 62.20 (61.00) & 37.50 (34.70) \\ \cmidrule(lr){2-6}
& Dogs (Easy)       & 100.0 (100.0) & 100.0 (70.33) & 100.0 (100.0) & 99.83 (93.50) \\
& Dogs (Medium)     & 93.50 (93.83) & 93.00 (64.83) & 86.67 (89.83) & 64.50 (50.33) \\
& Dogs (Hard)       & 64.00 (59.83) & 58.00 (43.50) & 56.83 (55.67) & 34.67 (33.00) \\ \cmidrule(lr){2-6}
& Food (Easy)       & 100.0 (99.80) & 100.0 (76.24) & 100.0 (100.0) & 99.80 (88.32) \\
& Food (Medium)     & 97.62 (97.03) & 98.02 (80.99) & 98.02 (97.82) & 91.09 (78.02) \\
& Food (Hard)       & 87.33 (84.95) & 81.39 (63.56) & 80.59 (84.75) & 48.91 (50.30) \\ \cmidrule(lr){2-6}
& iNaturalist (Easy)       & 95.75 (93.95) & 94.64 (80.50) & 96.31 (95.69) &  90.95 (68.79) \\
& iNaturalist (Medium)     & 83.62 (81.76) & 82.21 (65.33) & 81.06 (80.21) & 69.30 (56.87) \\
& iNaturalist (Hard)       & 41.84 (40.20) & 37.54 (29.88) & 37.89 (37.74) & 25.04 (25.84) \\

\bottomrule
\end{tabular}
\caption{Accuracy (\%) comparison of LVLMs across different dataset difficulty levels, with and without class names included in the option descriptions. Values in parentheses correspond to results under the "DCBA" option ordering; all others use the standard "ABCD" format.}
\label{tab:model_comparison}
\vspace{-0.5cm}
\end{table*}

%%%%%%%%%%%%%%%%%%%%%%%%%%%%%%%%%%%%%%%%%%%%%%%%%%%%%%%
\section{Selection Bias Mitigation}
\label{sec:debiasing}

To mitigate selection biases (positional and token) in LVLMs during MCQA tasks, we propose a bias mitigation framework comprising two main components: (1) ensemble bias vector estimation and (2) adaptive logit correction at inference time.

\paragraph{Ensemble Bias Vector Estimation.}
LVLMs often exhibit preferences toward certain answer tokens (e.g., "A", "B", "C", "D") due to training artifacts or prompt structures. To capture and correct for these biases, we estimate two types of bias vectors: general bias and contextual bias.

\textbf{General Bias Vector.}
The general bias vector \( B_{\text{gen}} \in \mathbb{R}^4 \) captures systematic biases that arise from model architecture, token position, or prompt format. It is computed by prompting the model with multiple semantically empty templates \( \{Q_1, Q_2, \dots, Q_n\} \), where each question has randomized answer option orders but lacks meaningful content. The model's output logits for each prompt are converted into probability distributions using softmax, and the results are averaged:
\begin{equation}
    B_{\text{gen}} = \frac{1}{n} \sum_{i=1}^n \text{softmax}(f_\theta(Q_i)).
\end{equation}
This estimates structural bias independent of task semantics, content, or reasoning cues.

\textbf{Contextual Bias Vector.}
The contextual bias vector \( B_{\text{ctx}} \in \mathbb{R}^4 \) measures how the model's predictions are skewed on a representative sample of actual data. Given a small (10\%), randomly selected subset of real MCQ examples \( \{(Q_j, A_j)\}_{j=1}^m \), we take the model's output logits and average the resulting probability distributions:
\begin{equation}
    B_{\text{ctx}} = \frac{1}{m} \sum_{j=1}^m \text{softmax}(f_\theta(Q_j)).
\end{equation}
This captures biases conditioned on realistic visual and textual inputs. Before combining, each bias vector is zero-centered by subtracting its mean to redistribute probabilities without introducing new preference directions.

\textbf{Final Ensemble Bias Vector.}
The two components are averaged to produce the final ensemble bias vector:
\begin{equation}
    B = \frac{B_{\text{gen}} + B_{\text{ctx}}}{2}.
\end{equation}
\noindent
\paragraph{Adaptive Logit Correction.}
During inference, the model outputs a logit vector \( L \in \mathbb{R}^4 \) for the four answer choices. We debias these logits using the ensemble bias vector:
\begin{equation}
    L' = L - \alpha_{\text{adp}} \cdot B,
\end{equation}
where \( \alpha_{\text{adp}} \) is a confidence-adaptive scaling factor defined as:\vspace{-0.2cm}
\begin{equation}
    \alpha_{\text{adp}} = \frac{\alpha}{1 + \exp(\text{conf} - \tau)}.
\end{equation}
Here, \( \alpha \) is a global scaling hyperparameter (default 1.0), and \( \tau \) is a threshold for model confidence (default 2.0). The confidence is computed as:
\begin{equation}
    \text{conf} = \max(L) - \text{mean}(L).
\end{equation}

This adaptive debiasing applies stronger correction when the model is uncertain (low confidence) and more conservative adjustment when the model is confident, helping prevent over-correction while mitigating selection biases.

%%%%%%%%%%%%%%%%%%%%%%%%%%%%%%%%%%%%%%%%%%%%%%%%%%%%%%%%%%

\section{Experiments and Evaluation}

In this section, we evaluate the extent of MCQA selection bias in LVLMs and assess the effectiveness of our proposed selection bias mitigation method. We benchmark model performance across different difficulty levels using our curated datasets with Qwen2.5-VL-3B-Instruct and 7B-Instruct (referred to as Qwen2.5-VL-3B or Qwen2.5-VL-7B throughout this paper, respectively), InternVL2\_5-8B, and LLaVA-v1.5-13B. We use 5 images per class for experiments at each difficulty level. For example, in the CUB "without class name" easy category, there are 200 classes, resulting in a total of 1,000 (200 $\times$ 5) images for evaluation.

%%%%%%%%%%%%%%%%%%%%%%%%%%%%%%%%%%%%%%%%%%%%%%%%%%% 

\subsection{LVLMs' Performance on MCQA Tasks}

Table~\ref{tab:model_comparison} presents a comprehensive comparison of LVLM performance on our curated fine-grained MCQA benchmarks, covering six datasets across three difficulty levels, with and without class names included in the option descriptions. We evaluate four prominent LVLM models: Qwen2.5-VL-3B-Instruct, Qwen2.5-VL-7B-Instruct, InternVL2\_5-8B, and LLaVA-v1.5-13B.

As expected, all models achieve near-perfect accuracy on easy tasks, confirming their ability to solve non-ambiguous MCQs regardless of format, though LLaVA-v1.5-13B shows notably lower performance on the iNaturalist dataset. However, performance diverges substantially on medium and hard tasks, particularly when class names are excluded. Qwen2.5-VL-7B-Instruct consistently outperforms others across all datasets and difficulty levels, demonstrating robustness under increasingly fine-grained conditions. InternVL2\_5-8B also performs competitively, especially on medium tasks. LLaVA-v1.5-13B, while strong on easy tasks, shows a notable accuracy drop on hard tasks, where the options are semantically fine-grained.

The inclusion of class names generally improves performance across all models, though to varying degrees. Additionally, results in the reversed option format (DCBA), shown in parentheses, indicate that certain models, most notably LLaVA-v1.5-13B, are more sensitive to selection biases. These findings underscore the importance of evaluating both content and structural factors in fine-grained MCQA tasks and highlight the varying degrees of bias susceptibility across LVLM architectures.

%%%%%%%%%%%%%%%%%%%%%%%%%%%%%%%%%%%%%%%%%%%%%%%%%%
\begin{table}[!ht]
\centering
\small
\begin{tabular}{llcc}
\toprule
\makecell{\textbf{Class} \\ \textbf{Name}} & 
\makecell{\textbf{Dataset} \\ \textbf{(Difficulty)}} & 
\makecell{\textbf{Qwen2.5-VL} \\ \textbf{-7B}} & 
\makecell{\textbf{LLaVA-v1.5} \\ \textbf{-13B}} \\
\midrule

\multirow{12}{*}{\textbf{With}} 
& Aircr. (M.) & --  & 67.71 (+18.86) \\
& Aircr. (H.)   & 71.71 (+0.57) & 21.71 (+00.00) \\ \cmidrule(lr){2-4}

& Cars (M.)      & -- & 93.47 (+8.78) \\
& Cars (H.)        & 83.67 (+1.53)  & 35.31 (+5.41) \\ \cmidrule(lr){2-4}

& CUB (M.)      & -- & 69.50 (+6.80) \\
& CUB (H.)        & 76.00 (-0.70)  & 30.00(+1.70) \\ \cmidrule(lr){2-4}

& Dogs (M.)     & -- & 66.33 (+21.33) \\
& Dogs (H.)       & 78.67 (+1.84) &  34.50 (+5.33) \\ \cmidrule(lr){2-4}

& Food (M.)     & -- & 92.67 (+2.77) \\
& Food (H.)       & 89.70 (+0.59) & 60.99 (+6.53)\\ 
\midrule

\multirow{12}{*}{\textbf{W/o}} 
& Aircr. (M.) & 84.00 (-0.57) & 48.86 (+10.29) \\
& Aircr. (H.)   & 54.86 (+1.15)  & 42.29 (+3.43)\\ \cmidrule(lr){2-4}

& Cars (M.)      & -- & 87.04 (+2.55) \\
& Cars (H.)        & 70.82 (+0.82) & 42.24 (+1.83) \\ \cmidrule(lr){2-4}

& CUB (M.)      & -- & 93.50 (+1.70) \\
& CUB (H.)        &  65.60 (+0.70)  & 38.60 (+1.10) \\ \cmidrule(lr){2-4}

& Dogs (M.)     & 94.50 (+1.00)  & 72.00 (+7.50) \\
& Dogs (H.)       & 64.33 (+0.33)  & 37.17 (+2.50) \\ \cmidrule(lr){2-4}

& Food (M.)     & 97.62 (+0.00)  & 95.84 (+4.75) \\
& Food (H.)       & 89.11 (+1.78) &51.88 (+2.97) \\ 

\bottomrule
\end{tabular}
\caption{Accuracy (\%) of our proposed selection bias mitigation method across multiple datasets and difficulty levels. Results are shown for Qwen2.5-VL-7B-Instruct and LLaVA-v1.5-13B under both class name ("With") and without class name ("W/o") settings. Numbers in parentheses indicate absolute gains over the standard “ABCD” format baseline. “--” denotes settings with near-saturated baseline accuracy ($\geq$98\%) where further improvement is not meaningful. M. and H. denote medium and hard difficulty, respectively. Aircr. represents FGVC Aircraft dataset.} 
\label{tab:bias_mitigation}
\vspace{-0.5cm}
\end{table}

%%%%%%%%%%%%%%%%%%%%%%%%%%%%%%%%%%%%%%%%%%%%%%%%%%%%%%%

\subsection{Bias Mitigation Results}

Table~\ref{tab:bias_mitigation} reports the performance of our proposed selection bias mitigation method on our curated fine-grained MCQA benchmarks: FGVC Aircraft, Stanford Cars, CUB, Stanford Dogs, and Food-101, under medium (M.) and hard (H.) difficulty levels. Results are shown for Qwen2.5-VL-7B-Instruct and LLaVA-v1.5-13B in both settings: with and without class names included in the option descriptions of the MCQs.

Since most LVLMs achieve 100\% accuracy on easy tasks in our benchmarks across all MCQA formats (e.g., ABCD, DCBA), we omit those cases from Table~\ref{tab:bias_mitigation}. However, we verify that our mitigation method does not degrade performance in these cases. For instance, on CUB with class name using LLaVA-v1.5-13B, the model retains 100\% accuracy even after applying mitigation.

We focus on Qwen2.5-VL-7B-Instruct and LLaVA-v1.5-13B as representative models due to their contrasting baseline performance: Qwen2.5-VL-7B achieves the highest accuracy overall, while LLaVA-v1.5-13B performs relatively poorly. This contrast enables us to evaluate the robustness and generalizability of our mitigation strategy across both strong and weak model conditions.

Our method consistently improves accuracy in most medium and hard settings, particularly under the "without class name" (W/o) condition, where selection biases are more pronounced. LLaVA-v1.5-13B exhibits substantial gains, such as +21.33\% on Dogs (Medium) and +6.53\% on Food (Hard). Qwen2.5-VL-7B-Instruct also shows improvements, including +1.15\% on Aircraft (Hard) and +1.84\% on Dogs (Hard). In medium-difficulty cases where baseline accuracy is already near-saturated ($\geq$98\%), gains are negligible or omitted. Only a few settings show minor accuracy drops (e.g., CUB (Hard task) with class names on Qwen2.5-VL-7B-Instruct), suggesting model- and dataset-specific variability.

Overall, these results demonstrate that our logit-based mitigation method effectively reduces selection bias without compromising accuracy, and often significantly improves it in challenging scenarios across our curated benchmarks.
%%%%%%%%%%%%%%%%%%%%%%%%%%%%%%%%%%%%%%%%%%%%%%%%%%%%%%%

\begin{table}[!ht]
    \small
    \centering
    \begin{tabular}{ccccc} \toprule
    \textbf{\makecell{Class \\Name}} & \textbf{Dataset}   & \textbf{Model} & \textbf{\makecell{ABCD \\(DCBA)}} & \textbf{\makecell{1234 \\(4321)}} \\ \midrule
    \multirow{9}{*}{With} & \multirow{3}{*}{CUB}	& LLaVA-13B	& \makecell{28.30 \\(33.70)}	& \makecell{27.20 \\(29.00)} \\ \cmidrule(lr){3-5}
    & & Qwen2.5-7B	& \makecell{76.70 \\(72.10)}	& \makecell{75.60 \\(66.80)} \\ \cmidrule(lr){2-5}
    & \multirow{3}{*}{Cars}	& LLaVA-13B	& \makecell{29.90 \\(35.51)}	& \makecell{25.20 \\(25.71)} \\ \cmidrule(lr){3-5}
    & & Qwen2.5-7B	& \makecell{82.14 \\(80.51)}	& \makecell{82.65 \\(78.98)} \\ \midrule
    
    \multirow{9}{*}{W/o} & \multirow{3}{*}{CUB}	& LLaVA-13B	& \makecell{37.50 \\(34.70)}	& \makecell{27.00 \\(27.10)} \\ \cmidrule(lr){3-5}
    & & Qwen2.5-7B	& \makecell{64.90 \\(63.30)}	& \makecell{64.60 \\(59.90)} \\ \cmidrule(lr){2-5}
    
    & \multirow{3}{*}{Cars}	& LLaVA-13B	& \makecell{40.41 \\(35.61)}	& \makecell{26.12 \\(26.73)} \\ \cmidrule(lr){3-5}
    & & Qwen2.5-7B	& \makecell{70.00 \\(67.86)}	& \makecell{65.92 \\(62.86)} \\ \bottomrule
    \end{tabular}
    \vspace{-0.2cm}
    \caption{Accuracy (\%) comparison between alphabetic (ABCD/DCBA) and numeric (1234/4321) option identifiers across different models and datasets for hard tasks. Parentheses show results with reversed order. For some datasets, we observe performance degradation with numeric option identifiers.}
    \label{tab:1234_result}
    \vspace{-0.4cm}
\end{table}

\subsection{Alternative Option IDs} \label{alt_option_ids}
We investigate alternative option identifiers to provide crucial insights into whether biases stem from specific token identity versus structural positioning effects. Our experiments with numeric identifiers (1/2/3/4), shown in Table~\ref{tab:1234_result}, reveal that biases persist regardless of the identifier type. For example, LLaVA-v1.5-13B exhibits a token bias toward `A' in ABCD and DCBA formats, while in 1234 and 4321 formats, it prefers option `1', indicating consistent token-level bias. Qwen2.5-VL-7B-Instruct similarly displays both token and position biases across all formats. Notably, performance drops significantly when using numeric formats (1234/4321). These findings suggest that LVLMs exhibit systematic preferences influenced by both token identity and structural positioning in the prompt format.

%%%%%%%%%%%%%%%%%%%%%%%%%%%%%%%%%%%%%%%%%%%%%%%%%%%%%%%

\subsection{Generalizability of the Mitigation Method}

We assess the generalizability of our proposed bias mitigation method beyond templated formats to strengthen its robustness. Specifically, we use the Qwen3-32B~\cite{qwen3} language model to regenerate questions and options in a more natural MCQ style, given the original question, options, and correct answer. For these experiments, we use the CUB (without class name) and Dogs (with class name) datasets. Unlike our main benchmarks, we do not explicitly control the difficulty level in these regenerated samples. However, the experimental results in Table~\ref{tab:generalizability} indicate a comparable level of difficulty. Below is an example of a regenerated MCQ from the CUB (without class name) dataset for the “Black-footed Albatross” class.

\begin{tcolorbox}[colback=blue!5!white, colframe=blue!75!black, title=Multiple Choice Question]
\textbf{Question:} What color is the majority of the bird's body in the image?
\begin{itemize}[noitemsep]
  \item[A.] Dark brown
  \item[B.] Yellow
  \item[C.] Blue
  \item[D.] Red
\end{itemize}

\textbf{Correct Answer:} \textbf{A}
\end{tcolorbox}

For these experiments, we use the LLaVA-v1.5-13B model and report accuracy in the standard ``ABCD'' format. As shown in Table~\ref{tab:generalizability}, our proposed mitigation method effectively reduces selection bias on newly generated benchmarks that do not follow templated MCQA formats, yielding performance gains of 1.17–5.00\%. These results suggest that the mitigation approach addresses underlying model behaviors that persist independently of specific question templates.\vspace{-0.3cm}
\begin{table}[!ht]
    \small
    \centering
    \begin{tabular}{lcc} \\ \toprule
    \textbf{\makecell{Dataset \\ (Difficulty)}}   & \textbf{\makecell{Without \\ Mitigation}} & \textbf{\makecell{With \\ Mitigation}} \\ \midrule
    CUB w/o name (M.)	& 90.40	& 91.70 (+1.30)   \\ 
    CUB w/o name (H.)	& 49.50	& 54.50 (+5.00) \\
    Dogs with name (M.)	& 81.50	& 82.67 (+1.17) \\
    Dogs with name (H.)	& 45.83	& 48.00 (+2.17) \\ \bottomrule
    \end{tabular}
    %\vspace{-0.2cm}
    \caption{Accuracy (\%) of the LLaVA-v1.5-13B model on beyond-template MCQA tasks, with and without the proposed bias mitigation method. Results show consistent improvements across all datasets, with larger gains on hard (H.) tasks compared to medium (M.) tasks.}
    \label{tab:generalizability}
    \vspace{-0.3cm}
\end{table}

We deliberately adopt a templated format (“Which description matches this object?”) to isolate selection bias under controlled conditions. By standardizing the question structure while varying distractor difficulty across diverse domains (birds, dogs, aircraft, cars, food), we can precisely measure token- and position-based biases without introducing confounding variables from question phrasing. Importantly, many real-world applications, such as educational assessments, standardized tests, and evaluation systems, employ structured MCQA formats, making our findings directly applicable.

%%%%%%%%%%%%%%%%%%%%%%%%%%%%%%%%%%%%%%%%%%%%%%%%%%%%%%%%

\section{Conclusion}
We present a systematic analysis of selection bias in Large Vision-Language Models (LVLMs) within the multiple-choice question answering (MCQA) setting. Our findings show that selection bias, driven by both token identity and positional preferences, intensifies with increasing task difficulty, particularly when MCQ options are fine-grained. While stronger models tend to exhibit reduced bias compared to weaker models, none are entirely immune, underscoring a fundamental limitation in current LVLM reasoning capabilities. To address this, we introduce a simple yet effective inference-time logit debiasing method that mitigates these biases without requiring model retraining. Our approach improves accuracy and consistency across varied input configurations and difficulty levels. This work highlights the importance of fine-grained, difficulty-aware benchmarks for revealing nuanced model behaviors and guiding future improvements. In future work, we aim to extend our framework to encompass a broader range of multimodal tasks.

%%%%%%%%%%%%%%%%%%%%%%%%%%%%%%%%%%%%%%%%%%%%%%%%%%%%%%%%%

\section{Limitations}
Our work specifically targets selection bias in MCQA, allowing focused investigation of these mechanisms while not addressing other bias forms such as modality imbalance or cultural bias. Our mitigation approach is designed for MCQA tasks rather than open-ended generation, enabling precise logit-level corrections for multiple-choice selection behavior. We curated comprehensive benchmarks within classification-type datasets to systematically analyze how selection biases manifest across diverse domains and difficulty levels. This targeted approach provides foundational insights into selection bias mechanisms in LVLMs. Future work will extend these findings to broader bias types and generative tasks, and will provide more comprehensive analysis of model-specific attention patterns across different architectures..

%%%%%%%%%%%%%%%%%%%%%%%%%%%%%%%%%%%%%%%%%%%%%%%%%%%%%%%%%

\section{Acknowledgments}
We acknowledge Advanced Research Computing (ARC) at Virginia Tech for providing the computational resources and technical support that contributed to the results reported in this paper. We thank the reviewers for their constructive feedback, which helped improve this paper.

%\bibliography{custom}

\begin{thebibliography}{44}
\providecommand{\natexlab}[1]{#1}

\bibitem[{Adila et~al.(2024)Adila, Zhang, Han, and Wang}]{adila2024discovering}
Dyah Adila, Shuai Zhang, Boran Han, and Yuyang Wang. 2024.
\newblock Discovering bias in latent space: an unsupervised debiasing approach.
\newblock In \emph{Proceedings of the 41st International Conference on Machine Learning}, pages 246--261.

\bibitem[{Atabuzzaman et~al.(2025)Atabuzzaman, Zhang, and Thomas}]{atabuzzaman2025}
Md~Atabuzzaman, Andrew Zhang, and Chris Thomas. 2025.
\newblock Zero-shot fine-grained image classification using large vision-language models.
\newblock In \emph{Findings of the Association for Computational Linguistics: EMNLP 2025}.

\bibitem[{Balepur et~al.(2025)Balepur, Rudinger, and Boyd-Graber}]{balepur2025these}
Nishant Balepur, Rachel Rudinger, and Jordan~Lee Boyd-Graber. 2025.
\newblock Which of these best describes multiple choice evaluation with llms? a) forced b) flawed c) fixable d) all of the above.
\newblock \emph{arXiv preprint arXiv:2502.14127}.

\bibitem[{Bossard et~al.(2014)Bossard, Guillaumin, and Van~Gool}]{food}
Lukas Bossard, Matthieu Guillaumin, and Luc Van~Gool. 2014.
\newblock Food-101--mining discriminative components with random forests.
\newblock In \emph{Computer vision--ECCV 2014: 13th European conference, zurich, Switzerland, September 6-12, 2014, proceedings, part VI 13}, pages 446--461. Springer.

\bibitem[{Chen et~al.(2024{\natexlab{a}})Chen, Cao, Zhang, and Lu}]{chen2024quantifying}
Meiqi Chen, Yixin Cao, Yan Zhang, and Chaochao Lu. 2024{\natexlab{a}}.
\newblock Quantifying and mitigating unimodal biases in multimodal large language models: A causal perspective.
\newblock In \emph{Findings of the Association for Computational Linguistics: EMNLP 2024}, pages 16449--16469.

\bibitem[{Chen et~al.(2024{\natexlab{b}})Chen, Guo, Li, Zhang, and Feng}]{chen2024mitigating}
Peng Chen, Xiao-Yu Guo, Yuan-Fang Li, Xiaowang Zhang, and Zhiyong Feng. 2024{\natexlab{b}}.
\newblock Mitigating language bias of lmms in social intelligence understanding with virtual counterfactual calibration.
\newblock In \emph{Proceedings of the 2024 Conference on Empirical Methods in Natural Language Processing}, pages 1300--1310.

\bibitem[{Chen et~al.(2025)Chen, Wu, Liu, Pan, Liu, Xie, Yu, and Ruan}]{januspro}
Xiaokang Chen, Zhiyu Wu, Xingchao Liu, Zizheng Pan, Wen Liu, Zhenda Xie, Xingkai Yu, and Chong Ruan. 2025.
\newblock Janus-pro: Unified multimodal understanding and generation with data and model scaling.
\newblock \emph{arXiv preprint arXiv:2501.17811}.

\bibitem[{Chen et~al.(2024{\natexlab{c}})Chen, Wang, Cao, Liu, Gao, Cui, Zhu, Ye, Tian, Liu et~al.}]{internvl2_5}
Zhe Chen, Weiyun Wang, Yue Cao, Yangzhou Liu, Zhangwei Gao, Erfei Cui, Jinguo Zhu, Shenglong Ye, Hao Tian, Zhaoyang Liu, and 1 others. 2024{\natexlab{c}}.
\newblock Expanding performance boundaries of open-source multimodal models with model, data, and test-time scaling.
\newblock \emph{arXiv preprint arXiv:2412.05271}.

\bibitem[{Cheng et~al.(2023)Cheng, Tian, Liu, Chen, Wang, Chen, and Zhang}]{cheng2023can}
Siyuan Cheng, Bozhong Tian, Qingbin Liu, Xi~Chen, Yongheng Wang, Huajun Chen, and Ningyu Zhang. 2023.
\newblock Can we edit multimodal large language models?
\newblock In \emph{Proceedings of the 2023 Conference on Empirical Methods in Natural Language Processing}, pages 13877--13888.

\bibitem[{Dai et~al.(2023)Dai, Li, Li, Tiong, Zhao, Wang, Li, Fung, and Hoi}]{instructblip}
Wenliang Dai, Junnan Li, Dongxu Li, Anthony Meng~Huat Tiong, Junqi Zhao, Weisheng Wang, Boyang Li, Pascale Fung, and Steven Hoi. 2023.
\newblock Instructblip: Towards general-purpose vision-language models with instruction tuning.
\newblock \emph{Advances in neural information processing systems}, 37.

\bibitem[{Deitke et~al.(2024)Deitke, Clark, Lee, Tripathi, Yang, Park, Salehi, Muennighoff, Lo, Soldaini et~al.}]{molmo}
Matt Deitke, Christopher Clark, Sangho Lee, Rohun Tripathi, Yue Yang, Jae~Sung Park, Mohammadreza Salehi, Niklas Muennighoff, Kyle Lo, Luca Soldaini, and 1 others. 2024.
\newblock Molmo and pixmo: Open weights and open data for state-of-the-art multimodal models.
\newblock \emph{arXiv preprint arXiv:2409.17146}.

\bibitem[{Di~Cecco et~al.(2021)Di~Cecco, Barve, Belitz, Stucky, Guralnick, and Hurlbert}]{di2021observing}
Grace~J Di~Cecco, Vijay Barve, Michael~W Belitz, Brian~J Stucky, Robert~P Guralnick, and Allen~H Hurlbert. 2021.
\newblock Observing the observers: How participants contribute data to inaturalist and implications for biodiversity science.
\newblock \emph{BioScience}, 71(11):1179--1188.

\bibitem[{Gokhale et~al.(2020)Gokhale, Banerjee, Baral, and Yang}]{gokhale2020mutant}
Tejas Gokhale, Pratyay Banerjee, Chitta Baral, and Yezhou Yang. 2020.
\newblock Mutant: A training paradigm for out-of-distribution generalization in visual question answering.
\newblock In \emph{Proceedings of the 2020 Conference on Empirical Methods in Natural Language Processing (EMNLP)}, pages 878--892.

\bibitem[{Guda et~al.(2025)Guda, Francis, Ashungafac, Joe-Wong, and Busogi}]{gudatiny}
Blessed Guda, Lawrence Francis, Gabrial~Zencha Ashungafac, Carlee Joe-Wong, and Moise Busogi. 2025.
\newblock Tiny: Rethinking selection bias in llms: Quantification and mitigation using efficient majority voting.
\newblock In \emph{ICLR Workshop: Quantify Uncertainty and Hallucination in Foundation Models: The Next Frontier in Reliable AI}.

\bibitem[{Khosla et~al.(2011)Khosla, Jayadevaprakash, Yao, and Li}]{dogs}
Aditya Khosla, Nityananda Jayadevaprakash, Bangpeng Yao, and Fei-Fei Li. 2011.
\newblock Novel dataset for fine-grained image categorization: Stanford dogs.
\newblock In \emph{Proc. CVPR workshop on fine-grained visual categorization (FGVC)}, volume~2.

\bibitem[{Kim and Ji(2024)}]{finer}
Jeonghwan Kim and Heng Ji. 2024.
\newblock Finer: Investigating and enhancing fine-grained visual concept recognition in large vision language models.
\newblock In \emph{Proceedings of the 2024 Conference on Empirical Methods in Natural Language Processing}, pages 6187--6207.

\bibitem[{Krause et~al.(2013)Krause, Stark, Deng, and Fei-Fei}]{cars}
Jonathan Krause, Michael Stark, Jia Deng, and Li~Fei-Fei. 2013.
\newblock 3d object representations for fine-grained categorization.
\newblock In \emph{Proceedings of the IEEE international conference on computer vision workshops}, pages 554--561.

\bibitem[{Li et~al.(2023)Li, Li, Savarese, and Hoi}]{blip2}
Junnan Li, Dongxu Li, Silvio Savarese, and Steven Hoi. 2023.
\newblock Blip-2: Bootstrapping language-image pre-training with frozen image encoders and large language models.
\newblock In \emph{International conference on machine learning}, pages 19730--19742. PMLR.

\bibitem[{Li et~al.(2024)Li, Li, Xiang, Liu, Deng, and Garcia}]{li2024can}
Wangyue Li, Liangzhi Li, Tong Xiang, Xiao Liu, Wei Deng, and Noa Garcia. 2024.
\newblock Can multiple-choice questions really be useful in detecting the abilities of llms?
\newblock In \emph{Proceedings of the 2024 Joint International Conference on Computational Linguistics, Language Resources and Evaluation (LREC-COLING 2024)}, pages 2819--2834.

\bibitem[{Liu et~al.(2024{\natexlab{a}})Liu, Li, Li, and Lee}]{llava2}
Haotian Liu, Chunyuan Li, Yuheng Li, and Yong~Jae Lee. 2024{\natexlab{a}}.
\newblock Improved baselines with visual instruction tuning.
\newblock In \emph{Proceedings of the IEEE/CVF Conference on Computer Vision and Pattern Recognition}, pages 26296--26306.

\bibitem[{Liu et~al.(2024{\natexlab{b}})Liu, Li, Wu, and Lee}]{llava}
Haotian Liu, Chunyuan Li, Qingyang Wu, and Yong~Jae Lee. 2024{\natexlab{b}}.
\newblock Visual instruction tuning.
\newblock \emph{Advances in neural information processing systems}, 36.

\bibitem[{Maji et~al.(2013)Maji, Rahtu, Kannala, Blaschko, and Vedaldi}]{aircrafts}
Subhransu Maji, Esa Rahtu, Juho Kannala, Matthew Blaschko, and Andrea Vedaldi. 2013.
\newblock Fine-grained visual classification of aircraft.
\newblock \emph{arXiv preprint arXiv:1306.5151}.

\bibitem[{OpenAI(2024)}]{gpt4o}
OpenAI. 2024.
\newblock Gpt-4o system card.
\newblock \emph{arXiv preprint arXiv:2410.21276}.

\bibitem[{Pezeshkpour and Hruschka(2024)}]{pezeshkpour2024large}
Pouya Pezeshkpour and Estevam Hruschka. 2024.
\newblock Large language models sensitivity to the order of options in multiple-choice questions.
\newblock In \emph{Findings of the Association for Computational Linguistics: NAACL 2024}, pages 2006--2017.

\bibitem[{Radford et~al.(2021)Radford, Kim, Hallacy, Ramesh, Goh, Agarwal, Sastry, Askell, Mishkin, Clark et~al.}]{clip}
Alec Radford, Jong~Wook Kim, Chris Hallacy, Aditya Ramesh, Gabriel Goh, Sandhini Agarwal, Girish Sastry, Amanda Askell, Pamela Mishkin, Jack Clark, and 1 others. 2021.
\newblock Learning transferable visual models from natural language supervision.
\newblock In \emph{International conference on machine learning}, pages 8748--8763. PMLR.

\bibitem[{Robinson and Wingate(2023)}]{robinsonleveraging}
Joshua Robinson and David Wingate. 2023.
\newblock Leveraging large language models for multiple choice question answering.
\newblock In \emph{The Eleventh International Conference on Learning Representations (ICLR)}.

\bibitem[{Shi et~al.(2024)Shi, Ma, Liang, Ma, and Vosoughi}]{shi2024judging}
Lin Shi, Chiyu Ma, Wenhua Liang, Weicheng Ma, and Soroush Vosoughi. 2024.
\newblock Judging the judges: A systematic investigation of position bias in pairwise comparative assessments by llms.
\newblock \emph{arXiv preprint arXiv:2406.07791}.

\bibitem[{Tan et~al.(2024)Tan, Chu, Li, and Mo}]{tan2024order}
Zhijie Tan, Xu~Chu, Weiping Li, and Tong Mo. 2024.
\newblock Order matters: Exploring order sensitivity in multimodal large language models.
\newblock \emph{arXiv preprint arXiv:2410.16983}.

\bibitem[{Team(2025)}]{Qwen2.5-VL}
Qwen Team. 2025.
\newblock \href {https://qwenlm.github.io/blog/qwen2.5-vl/} {Qwen2.5-vl}.

\bibitem[{Wah et~al.(2011)Wah, Branson, Welinder, Perona, and Belongie}]{cub}
Catherine Wah, Steve Branson, Peter Welinder, Pietro Perona, and Serge Belongie. 2011.
\newblock The caltech-ucsd birds-200-2011 dataset.
\newblock \emph{California Institute of Technology}.

\bibitem[{Wang et~al.(2025)Wang, Zhao, Qiang, Xi, Qin, and Liu}]{wang2025llms}
Haochun Wang, Sendong Zhao, Zewen Qiang, Nuwa Xi, Bing Qin, and Ting Liu. 2025.
\newblock Llms may perform mcqa by selecting the least incorrect option.
\newblock In \emph{Proceedings of the 31st International Conference on Computational Linguistics}, pages 5852--5862.

\bibitem[{Wang et~al.(2021)Wang, Liu, and Wang}]{wang2021gender}
Jialu Wang, Yang Liu, and Xin Wang. 2021.
\newblock Are gender-neutral queries really gender-neutral? mitigating gender bias in image search.
\newblock In \emph{Proceedings of the 2021 Conference on Empirical Methods in Natural Language Processing}, pages 1995--2008.

\bibitem[{Wang et~al.(2024{\natexlab{a}})Wang, Li, Chen, Cai, Zhu, Lin, Cao, Kong, Liu, Liu et~al.}]{wang2024large}
Peiyi Wang, Lei Li, Liang Chen, Zefan Cai, Dawei Zhu, Binghuai Lin, Yunbo Cao, Lingpeng Kong, Qi~Liu, Tianyu Liu, and 1 others. 2024{\natexlab{a}}.
\newblock Large language models are not fair evaluators.
\newblock In \emph{Proceedings of the 62nd Annual Meeting of the Association for Computational Linguistics (Volume 1: Long Papers)}, pages 9440--9450.

\bibitem[{Wang et~al.(2024{\natexlab{b}})Wang, Li, Qin, Li, Tu, Chu, and Sui}]{wang2024can}
Zecheng Wang, Xinye Li, Zhanyue Qin, Chunshan Li, Zhiying Tu, Dianhui Chu, and Dianbo Sui. 2024{\natexlab{b}}.
\newblock Can we debias multimodal large language models via model editing?
\newblock In \emph{Proceedings of the 32nd ACM International Conference on Multimedia}, pages 3219--3228.

\bibitem[{Wei et~al.(2024)Wei, Wu, Huang, and Chen}]{wei2024unveiling}
Sheng-Lun Wei, Cheng-Kuang Wu, Hen-Hsen Huang, and Hsin-Hsi Chen. 2024.
\newblock Unveiling selection biases: Exploring order and token sensitivity in large language models.
\newblock In \emph{Findings of the Association for Computational Linguistics ACL 2024}, pages 5598--5621.

\bibitem[{Xue et~al.(2024)Xue, Hu, Liu, Liao, Han, Zhao, Yin et~al.}]{xue2024strengthened}
Mengge Xue, Zhenyu Hu, Liqun Liu, Kuo Liao, Honglin Han, Meng Zhao, Chengguo Yin, and 1 others. 2024.
\newblock Strengthened symbol binding makes large language models reliable multiple-choice selectors.
\newblock In \emph{Proceedings of the 62nd Annual Meeting of the Association for Computational Linguistics (Volume 1: Long Papers)}, pages 4331--4344.

\bibitem[{Yang et~al.(2025{\natexlab{a}})Yang, Li, Yang, Zhang, Hui, Zheng, Yu, Chang, and so~on...}]{qwen3}
An~Yang, Anfeng Li, Baosong Yang, Beichen Zhang, Binyuan Hui, Bo~Zheng, Bowen Yu, Chang, and so~on... 2025{\natexlab{a}}.
\newblock Qwen3 technical report.
\newblock \emph{arXiv preprint arXiv:2505.09388}.

\bibitem[{Yang et~al.(2024)Yang, Kang, Choi, Lee, and Jung}]{yang2024mitigating}
Nakyeong Yang, Taegwan Kang, Stanley~Jungkyu Choi, Honglak Lee, and Kyomin Jung. 2024.
\newblock Mitigating biases for instruction-following language models via bias neurons elimination.
\newblock In \emph{Proceedings of the 62nd Annual Meeting of the Association for Computational Linguistics (Volume 1: Long Papers)}, pages 9061--9073.

\bibitem[{Yang et~al.(2025{\natexlab{b}})Yang, Jian, and Li}]{yang2025option}
Zhen Yang, Ping Jian, and Chengzhi Li. 2025{\natexlab{b}}.
\newblock Option symbol matters: Investigating and mitigating multiple-choice option symbol bias of large language models.
\newblock In \emph{Proceedings of the 2025 Conference of the Nations of the Americas Chapter of the Association for Computational Linguistics: Human Language Technologies (Volume 1: Long Papers)}, pages 1902--1917.

\bibitem[{Zhang et~al.(2024)Zhang, Yu, Wen, Wang, Zhang, Wang, Jin, and Tan}]{zhang2024debiasing}
Yi-Fan Zhang, Weichen Yu, Qingsong Wen, Xue Wang, Zhang Zhang, Liang Wang, Rong Jin, and Tieniu Tan. 2024.
\newblock Debiasing multimodal large language models.
\newblock \emph{arXiv preprint arXiv:2403.05262}.

\bibitem[{Zheng et~al.(2024)Zheng, Zhou, Meng, Zhou, and Huang}]{zheng2024large}
Chujie Zheng, Hao Zhou, Fandong Meng, Jie Zhou, and Minlie Huang. 2024.
\newblock Large language models are not robust multiple choice selectors.
\newblock In \emph{ICLR}.

\bibitem[{Zheng et~al.(2023)Zheng, Chiang, Sheng, Zhuang, Wu, Zhuang, Lin, Li, Li, Xing et~al.}]{zheng2023judging}
Lianmin Zheng, Wei-Lin Chiang, Ying Sheng, Siyuan Zhuang, Zhanghao Wu, Yonghao Zhuang, Zi~Lin, Zhuohan Li, Dacheng Li, Eric Xing, and 1 others. 2023.
\newblock Judging llm-as-a-judge with mt-bench and chatbot arena.
\newblock \emph{Advances in Neural Information Processing Systems}, 36:46595--46623.

\bibitem[{Zhou et~al.(2024)Zhou, Feng, Zhu, Qian, and Mao}]{zhou2024unibias}
Hanzhang Zhou, Zijian Feng, Zixiao Zhu, Junlang Qian, and Kezhi Mao. 2024.
\newblock Unibias: Unveiling and mitigating llm bias through internal attention and ffn manipulation.
\newblock In \emph{The Thirty-eighth Annual Conference on Neural Information Processing Systems}.

\bibitem[{Zong et~al.(2024)Zong, Yu, Chavhan, Zhao, and Hospedales}]{zong2024fool}
Yongshuo Zong, Tingyang Yu, Ruchika Chavhan, Bingchen Zhao, and Timothy Hospedales. 2024.
\newblock Fool your (vision and) language model with embarrassingly simple permutations.
\newblock In \emph{International Conference on Machine Learning}, pages 62892--62913. PMLR.

\end{thebibliography}

\appendix

\section{Appendix}
\label{sec:appendix}

This section contains the following topics.
\begin{itemize}[noitemsep]
    \item Dataset Statistics (Appendix~\ref{dataset_sta})
    \item Visualization of LVLMs' Selection Bias (Appendix~\ref{vis_select})
\end{itemize}

%------------------------------------------

\subsection{Dataset Statistics}
\label{dataset_sta}

To construct each MCQ, we utilize existing fine-grained image classification datasets, including CUB, Stanford Dogs, FGVC Aircraft, Stanford Cars, Food-101, and iNaturalist. Each image is paired with a curated class description. Class descriptions are collected from \citet{atabuzzaman2025}. For the iNaturalist dataset, we collect class descriptions from \citet{finer} and use common names of species instead of scientific class names. Upon analysis, we found 38 species with duplicate common names, resulting in 9,962 unique classes.

We analyze the datasets by reporting the average and standard deviation of semantic similarity scores across the three difficulty levels in Table \ref{tab:similarity_stats}. As expected, the Easy set exhibits low average similarity, the Medium set falls in the mid-range, and the Hard set shows high similarity between the target and distractors. These statistics validate our difficulty categorization and provide a quantitative basis for evaluating model performance across different levels of semantic and visual ambiguity.

\begin{table*}[ht]
\centering
\small
\begin{tabular}{llcccc}
\toprule
\textbf{Dataset} & \textbf{Class Names} & \textbf{Difficulty} & \textbf{Domain} & \textbf{Avg. Sim.} & \textbf{Std. Dev.} \\
\midrule

\multirow{6}{*}{CUB}
  & \multirow{3}{*}{Without} & Easy   & Diff. & 0.2241 & 0.0409 \\
  &                          & Medium & Same  & 0.4259 & 0.0472 \\
  &                          & Hard   & Same  & 0.8461 & 0.0573 \\
  \cmidrule(lr){2-6}
  & \multirow{3}{*}{With}    & Easy   & Diff. & 0.1864 & 0.0408 \\
  &                          & Medium & Same  & 0.3696 & 0.0347 \\
  &                          & Hard   & Same  & 0.7831 & 0.0707 \\

\midrule

\multirow{6}{*}{Dogs}
  & \multirow{3}{*}{Without} & Easy   & Diff. & 0.1736 & 0.0311 \\
  &                          & Medium & Same  & 0.5574 & 0.0557 \\
  &                          & Hard   & Same  & 0.8855 & 0.0536 \\
  \cmidrule(lr){2-6}
  & \multirow{3}{*}{With}    & Easy   & Diff. & 0.1382 & 0.0354 \\
  &                          & Medium & Same  & 0.3854 & 0.0573 \\
  &                          & Hard   & Same  & 0.7981 & 0.0546 \\

\midrule

\multirow{6}{*}{Aircraft}
  & \multirow{3}{*}{Without} & Easy   & Diff. & 0.1478 & 0.0427 \\
  &                          & Medium & Same  & 0.6971 & 0.0509 \\
  &                          & Hard   & Same  & 0.9413 & 0.0369 \\
  \cmidrule(lr){2-6}
  & \multirow{3}{*}{With}    & Easy   & Diff. & 0.1623 & 0.0397 \\
  &                          & Medium & Same  & 0.4875 & 0.0531 \\
  &                          & Hard   & Same  & 0.8256 & 0.0437 \\

\midrule

\multirow{6}{*}{Cars}
  & \multirow{3}{*}{Without} & Easy   & Diff. & 0.1444 & 0.0366 \\
  &                          & Medium & Same  & 0.6458 & 0.0462 \\
  &                          & Hard   & Same  & 0.9203 & 0.0356 \\
  \cmidrule(lr){2-6}
  & \multirow{3}{*}{With}    & Easy   & Diff. & 0.1350 & 0.0357 \\
  &                          & Medium & Same  & 0.3571 & 0.0502 \\
  &                          & Hard   & Same  & 0.8237 & 0.0732 \\

\midrule

\multirow{6}{*}{Food}
  & \multirow{3}{*}{Without} & Easy   & Diff. & 0.1969 & 0.0402 \\
  &                          & Medium & Same  & 0.3228 & 0.0529 \\
  &                          & Hard   & Same  & 0.6942 & 0.0615 \\
  \cmidrule(lr){2-6}
  & \multirow{3}{*}{With}    & Easy   & Diff. & 0.1873 & 0.0377 \\
  &                          & Medium & Same  & 0.3856 & 0.0434 \\
  &                          & Hard   & Same  & 0.7066 & 0.0532 \\

\midrule

\multirow{6}{*}{iNaturalist}
  & \multirow{3}{*}{Without} & Easy   & Diff. & 0.1308 & 0.0490 \\
  &                          & Medium & Same  & 0.3438 & 0.1173 \\
  &                          & Hard   & Same  & 0.8408 & 0.0708 \\
  \cmidrule(lr){2-6}
  & \multirow{3}{*}{With}    & Easy   & Diff. & 0.0888 & 0.0334 \\
  &                          & Medium & Same  & 0.2844 & 0.1431 \\
  &                          & Hard   & Same  & 0.7813 & 0.0818 \\

\bottomrule
\end{tabular}
\caption{Average similarity and standard deviation between the ground truth and distractor options across difficulty levels (Easy, Medium, Hard) on fine-grained datasets, with and without class names. Easy distractors are from different domains; Medium and Hard are from the same domain. Std. is computed among the distractors.}
\label{tab:similarity_stats}
\end{table*}

\paragraph{Answer Option Distribution.} Table~\ref{tab:answer_distribution} presents the distribution of correct answer positions (A–D) across our benchmark datasets, segmented by dataset (CUB, FGVC Aircraft, Stanford Cars, Stanford Dogs, Food-101, and iNaturalist), difficulty level (Easy, Medium, Hard), and the presence or absence of class names in the answer options. Each row corresponds to a specific configuration, and the columns report how many times each option ID (A, B, C, D) is the correct answer.

To ensure fair evaluation of selection bias, we constructed all datasets to maintain a near-balanced distribution of correct answers across the four option IDs. This prevents any skew that could arise from inherent imbalances in the dataset and helps isolate model behavior due to selection bias rather than label distribution.

Each difficulty level contains both "with class name" and "without class name" variants, allowing us to assess the role of explicit label cues in model predictions. For example, in the CUB (Easy) category, the correct answer is evenly distributed across the four choices (A–D) for both versions. Similar patterns are preserved across other datasets and difficulty levels.

Across all datasets and configurations, the total number of multiple-choice questions is 63,894, with each variant carefully designed to mitigate structural bias in the ground-truth distribution. This careful balancing enables controlled investigation of the positional and token-level biases of LVLMs during multiple-choice question answering.

\begin{table*}[htbp]
\small 
\centering
\begin{tabular}{llccccc}
\toprule
\multirow{2}{*}{\textbf{Category}} & \multirow{2}{*}{\textbf{Class Name}} & \multicolumn{4}{c}{\textbf{Option IDs}} &\multirow{2}{*}{\textbf{Total}} \\
\cmidrule{3-6}
& &\textbf{A} & \textbf{B} & \textbf{C} & \textbf{D} & \\ \midrule
\multirow{2}{*}{CUB (Easy)}       & Without & 42 & 54 & 54 & 50 & 200 \\
                                  & With    & 42 & 54 & 54 & 50 & 200 \\
\midrule
\multirow{2}{*}{CUB (Medium+Hard)}  & Without &112  & 90 & 94 & 104  & 400 \\ 
                                & With & 100 & 95 & 103 & 102 & 400 
 \\ 
\midrule 

\multirow{2}{*}{Aircraft (Easy)} & Without & 21 & 20 & 18 & 11 & 70 \\
                                  & With    & 21 & 20 & 18 & 11 & 70 \\
\midrule
\multirow{2}{*}{Aircraft (Medium+Hard)}  & Without & 36  & 41  & 25 & 38  & 140 \\ 
                                & With & 28  & 43  &37  & 32  & 140 \\ 
\midrule 
\multirow{2}{*}{Cars (Easy)}      & Without & 56 & 54 & 38 & 48 & 196 \\
                                  & With    & 56 & 54 & 38 & 48 & 196 \\
\midrule
\multirow{2}{*}{Cars (Medium+Hard)}  & Without &85  & 99 & 108 & 100 & 392 \\ 
                                & With & 92  & 106  & 97 & 97  & 392  \\ 
\midrule 

\multirow{2}{*}{Dogs (Easy)}      & Without & 28 & 25 & 28 & 39 & 120 \\
                                  & With    & 28 & 25 & 28 & 39 & 120 \\
\midrule
\multirow{2}{*}{Dogs (Medium+Hard)}  & Without & 72 & 54 & 57 & 57 & 240 \\ 
                                & With & 60  & 71  & 59  & 50 & 240  \\ 
                                \hline 
\multirow{2}{*}{Food (Easy)}     & Without & 24 & 26 & 27 & 24 & 101 \\
                                  & With    & 24 & 26 & 27 & 24 & 101 \\
\midrule
\multirow{2}{*}{Food (Medium+Hard)}  & Without & 53 & 48 & 52 & 49 & 202  \\ 
                                & With & 50  & 49  & 49 & 54  & 202 \\ 
                                \hline 
\multirow{2}{*}{iNaturalist (Easy)} & Without & 2491 & 2474 & 2439 & 2558 & 9962 \\
                                    & With    & 2491 & 2474 & 2439 & 2558  & 9962 \\
\midrule
\multirow{2}{*}{iNaturalist (Medium)}  & Without & 2545 & 2505 & 2417 & 2495 & 9962 \\ 
                                & With & 2498 & 2533 & 2430 & 2501 & 9962 \\ 
\midrule 
\multirow{2}{*}{iNaturalist (Hard)}  & Without & 2469 & 2508 & 2453 & 2532 & 9962 \\ 
                                & With & 2469 & 2508 & 2453 & 2532 & 9962 \\ 
\midrule 
\multicolumn{7}{r}{\textbf{Total = 63894}} \\
\bottomrule
\end{tabular}
\caption{Distribution of correct answer options (A–D) across datasets for Easy, Medium, and Hard categories. For brevity, we merge Medium and Hard into a single row for smaller datasets. Each dataset maintains a balanced distribution of answer options across all categories, both with and without the class name.}
\label{tab:answer_distribution}
\end{table*}

\subsection{Visualization of LVLMs' Selection Bias}
\label{vis_select}

\begin{figure*}[!ht]
    \centering
        \begin{subfigure}[b]{0.95\linewidth}
        \centering
        \includegraphics[width=\textwidth]{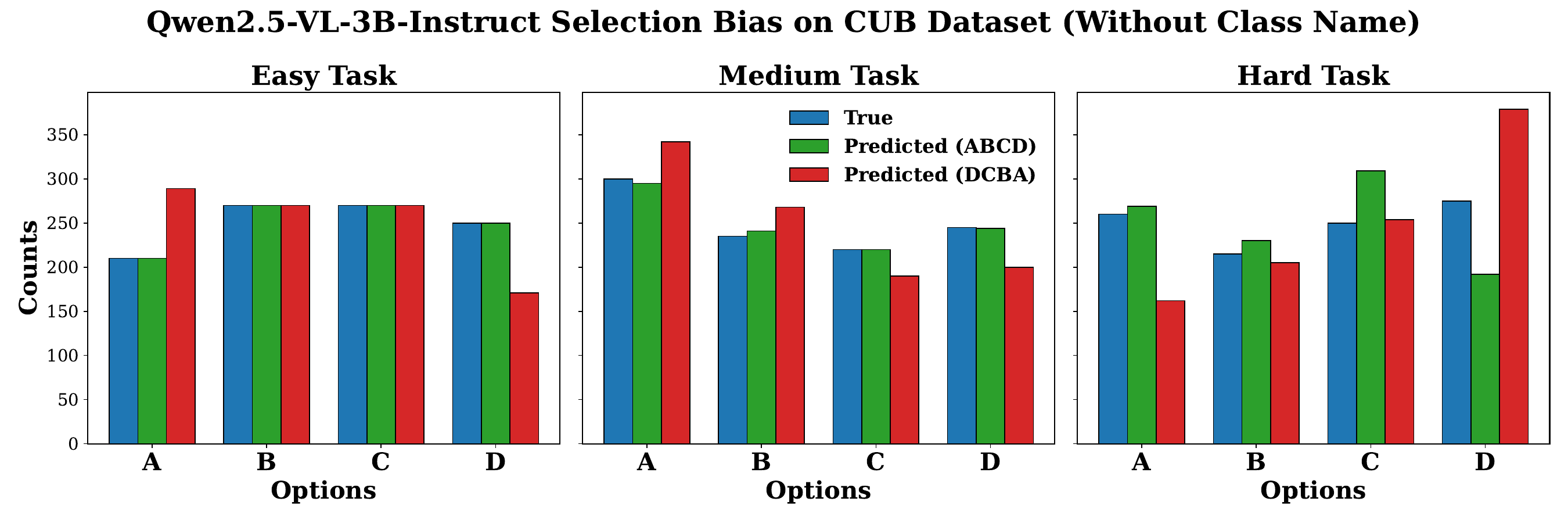}
        \caption{ Qwen2.5-VL-3B-Instruct exhibits stronger selection biases—favoring option ID “A” in Easy tasks (\textbf{token bias}) and the first-position options “A” and “D” in “ABCD” and “DCBA” orderings, respectively, in Hard tasks (\textbf{positional bias}).}
        \label{fig:qwen_3b_bias_app}
    \end{subfigure}
    \vskip\baselineskip
    \begin{subfigure}[b]{0.95\linewidth}
        \centering
        \includegraphics[width=\textwidth]{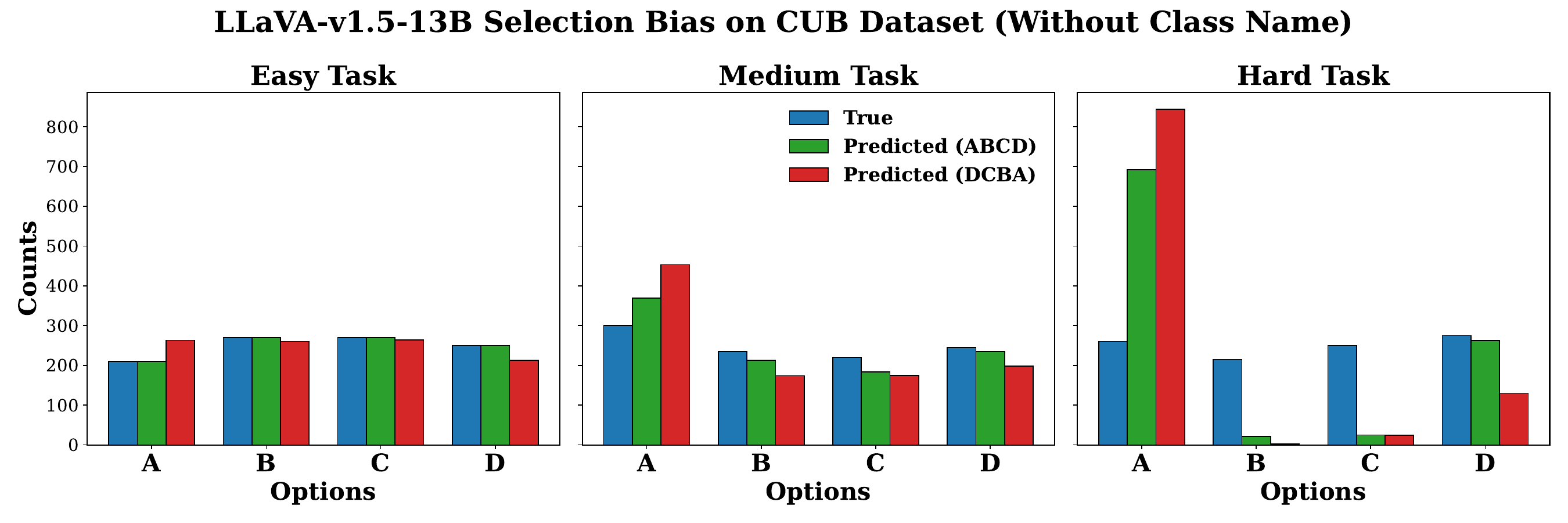}
        \caption{LLaVA-v1.5-13B model shows balanced selection in Easy tasks but develops strong token bias in Hard tasks, with dramatic preference for the option ID "A" (reaching nearly 3x the true frequency) when difficulty increases.}
        \label{fig:llava_13b_bias_app}
    \end{subfigure}
    \vskip\baselineskip
    \begin{subfigure}[b]{0.95\linewidth}
        \centering
        \includegraphics[width=\textwidth]{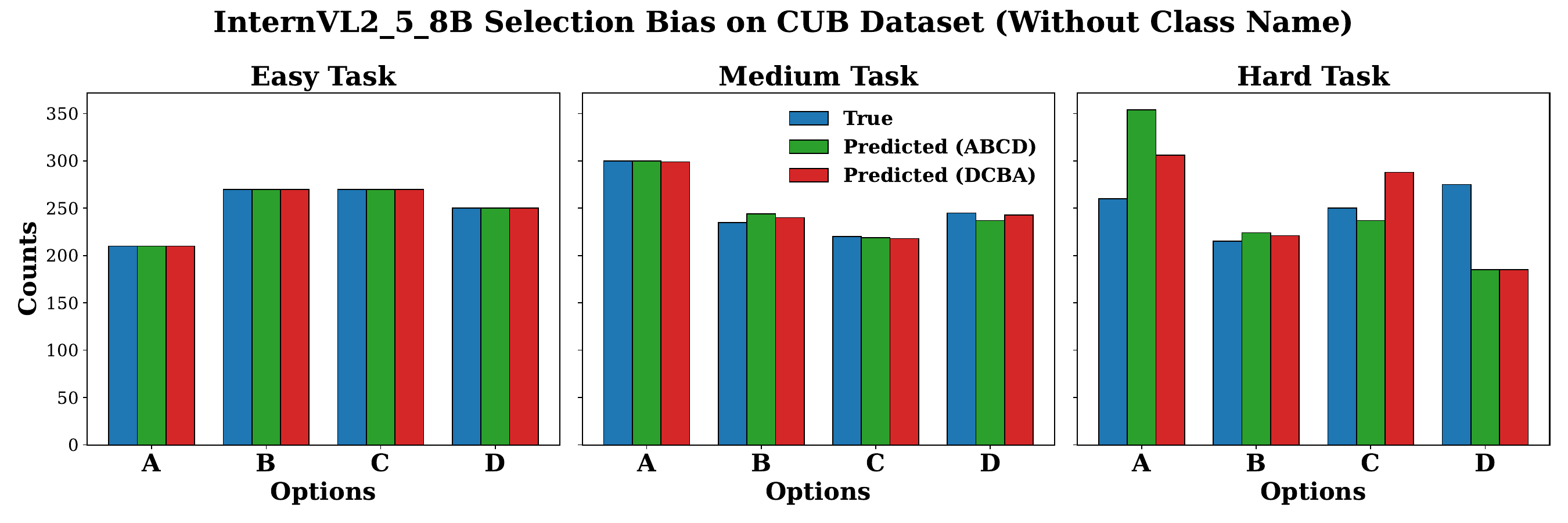}
        \caption{InternVL2\_5-8B model demonstrates the most balanced behavior across difficulty levels, with predictions closely matching the true distribution in Easy and Medium tasks. In Hard tasks, it shows moderate token bias ("A") but maintains better distribution consistency between ABCD and DCBA orderings.}
        \label{fig:internvl_8b_bias_app}
    \end{subfigure}
    \caption{Selection bias comparison across LVLMs on the CUB dataset under the "without class name" setting, organized by increasing task difficulty (Easy, Medium, Hard). Each plot shows distributions for ground truth (True) and model predictions under two option orderings: standard (ABCD) and reversed (DCBA). The comparison reveals how position and token biases emerge and intensify with task difficulty, with varying patterns across architectures.}
    \label{fig:all_models_bias_app}
    \vspace{-0.5cm}
\end{figure*}

Figure~\ref{fig:all_models_bias_app} provides a comparative visualization of selection bias exhibited by three state-of-the-art Large Vision-Language Models (LVLMs): Qwen2.5-VL-3B-Instruct, LLaVA-v1.5-13B, and InternVL2\_5-8B on the CUB dataset under the "without class name" setting. The bar plots illustrate how each model's answer distribution changes across task difficulty levels (Easy, Medium, Hard) and under different option orderings (standard ABCD vs. reversed DCBA). As the task becomes more difficult, Qwen2.5-VL-3B shows increasing bias toward specific token identities and positions, such as a consistent preference for option “A” or the first-position choice. The LLaVA-v1.5-13B model shows balanced selection in easy tasks but develops strong token bias in hard tasks, exhibiting a dramatic preference for option ID “A” (nearly three times the true frequency) as uncertainty increases. In contrast, InternVL2\_5-8B demonstrates more stable behavior across conditions, maintaining a distribution that closely aligns with the ground truth, especially in Medium tasks. These visualizations highlight both shared and model-specific patterns of bias, reinforcing the importance of task difficulty and option formatting when evaluating LVLM behavior.

\end{document}